\documentclass[]{ceurart}

\usepackage{listings}
\usepackage{tabularx}

\usepackage{pgfplots}
\usepackage{pgfplotstable}
\usepackage{booktabs}
\usepackage{array}
\usepackage{colortbl}

\usetikzlibrary{shapes.geometric, arrows.meta, positioning, fit}

\begin{document}

\copyrightyear{2024}
\copyrightclause{Copyright for this paper by its authors.
  Use permitted under Creative Commons License Attribution 4.0 International (CC BY 4.0).}

\conference{KBC-LM'24: Knowledge Base Construction from Pre-trained Language Models workshop at ISWC 2024}

\title{The Effects of Hallucinations in Synthetic Training Data for Relation Extraction}

\author[1]{Steven Rogulsky}[%
email=steven.rogulsky@student.kit.edu,
]
\address[1]{Karlsruhe Institute of Technology (KIT), Karlsruhe, Germany}

\author[1]{Nicholas Popovic}[%
email=popovic@kit.edu,
]

\author[2]{Michael Färber}[%
email=michael.faerber@tu-dresden.de,
]
\address[2]{TU Dresden \& ScaDS.AI, Dresden, Germany}

\begin{abstract}
Relation extraction is crucial for constructing knowledge graphs, with large high-quality datasets serving as the foundation for training, fine-tuning, and evaluating models. Generative data augmentation (GDA) is a common approach to expand such datasets. However, this approach often introduces hallucinations, such as spurious facts, whose impact on relation extraction remains underexplored. In this paper, we examine the effects of 
hallucinations on the performance of relation extraction on the document and sentence levels. Our empirical study reveals 
that hallucinations considerably compromise the ability of models to extract relations from text, with recall reductions between 19.1\% and 39.2\%. 
We identify that relevant hallucinations impair the model's performance, while irrelevant hallucinations have a minimal impact. 
Additionally, we develop methods for the detection of hallucinations to improve data quality and model performance. Our approaches successfully classify texts as either 'hallucinated' or 'clean,' achieving high F1-scores of 83.8\% and 92.2\%. These methods not only assist in removing hallucinations but also help in estimating their prevalence within datasets, which is crucial for selecting high-quality data. 
Overall, our work confirms the profound impact of relevant hallucinations on the effectiveness of relation extraction models. 
\end{abstract}

\maketitle

\section{Introduction}
\label{Introduction}

Relation extraction is an important step in extracting structured information from text documents, such as news articles, publications, patents, and websites, building the basis for knowledge graph construction. High-quality datasets play a crucial role in this process \cite{hernandez-garcia_data_2020, gao_self-guided_2023, wei_eda_2019}, as they form the basis for training, fine-tuning, and evaluating relation extraction models. %
Additionally, the amount of data they contain has a significant impact on the achieved results \cite{anaby-tavor_not_2019}. However, creating large datasets with high quality typically requires human annotation, which is expensive and slow. Although heuristic methods such as distant supervision can produce larger datasets, they often lack quality \cite{huguet_cabot_rebel_2021}. An alternative is \textit{Generative Data Augmentation (GDA)}, a technique for synthetically expanding datasets by generating new data samples (here: texts and extracted triples). It can generate 
datasets that are much larger, more diverse, and less expensive than traditional human annotations without directly collecting new data \cite{josifoski_exploiting_2023}.  
In the context of relation extraction, GDA has been widely used in combination with pre-trained language models such as BERT and GPT \cite{ribeiro_investigating_2021, wang_stage-wise_2021, chen_mixtext_2020, thakur_augmented_2021, yoo_gpt3mix_2021}.

\begin{figure}[tb]
    \centering
\noindent
\resizebox{\textwidth}{!}{
\begin{tikzpicture}[
    box/.style={rectangle, draw=black, rounded corners, minimum height=1cm, minimum width=6cm, align=center, text width=6cm, inner sep=10pt},
    compactbox/.style={rectangle, draw=black, rounded corners, minimum height=1cm, minimum width=3cm, align=center, text width=3cm, inner sep=5pt}, %
    emphgreen/.style={text=gray, font=\bfseries},
    emphred/.style={text=red, font=\bfseries},
    emphblue/.style={text=blue, font=\bfseries},
    arrow/.style={-Stealth, thick},
    rectarrow/.style={-Stealth, thick, to path={-| (\tikztotarget)}},
    outarrow/.style={-Stealth, thick, rounded corners}
]

\node[box, text width=5cm] (triple) {
    \textbf{Triple} \\ 
    (\textcolor{gray}{Ted}, \textcolor{red}{'lives in'}, \textcolor{blue}{'New York'})
};

\node[compactbox, right=1cm of triple] (model) {
    \textbf{GDA-Model}
};

\node[box, right=2cm of model, yshift=1cm, text width=8cm] (nohallucination) {
    \textbf{Text without Hallucinations} \\
    \textcolor{gray}{Ted} \textcolor{red}{lives in} the city of \textcolor{blue}{New York}
};

\node[box, right=2cm of model, yshift=-1cm, text width=8cm] (hallucination) {
    \textbf{Text with Hallucinations} \\
    \textcolor{gray}{Ted} \textcolor{red}{lives in} the city of \textcolor{blue}{New York}, which has a population of \textbf{\textcolor{blue}{8.4 million inhabitants}}.
};

\draw[arrow] (triple.east) -- (model.west); %

\draw[arrow] (model.east) -- ++(1,0) |- (nohallucination.west); %
\draw[arrow] (model.east) -- ++(1,0) |- (hallucination.west);   %

\end{tikzpicture}
}
\vspace{-0.5cm}
\caption{Text generated by Generative Data Augmentation (GDA) with and without hallucinations.}
    \label{fig:example_hallucination}
\end{figure}
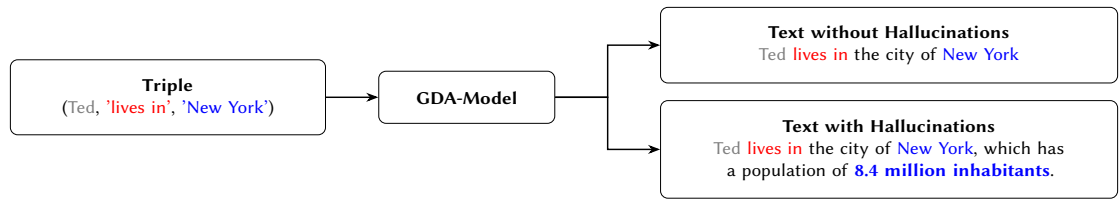

Despite its advantages, GDA often leads to \textit{hallucinations} in the text, where the content deviates from the information in the input, as, for instance, additional facts are generated (see Figure~1). This issue 
commonly occurs in generative language models~\cite{ye_cognitive_2023,ji_survey_2023,varshney_stitch_2023}. 
If a language model is trained on a dataset with incorrect annotations due to hallucinations, 
the effectiveness of the relation extraction method may be compromised -- to which degree is unknown --, potentially reducing the accuracy of extracted triples as the model might not learn to capture all necessary information. 
Although the phenomenon of hallucinations is well-recognized, and the use of LLMs to generate training datasets is increasing, the specific effects of 
hallucinations on relation extraction have not been thoroughly investigated. 

In this paper, we examine the impact of hallucinations in synthetic training data on relation extraction, considering several hallucination types on the document and sentence level. Our research focuses on two primary questions: 
\textbf{RQ1: Can we detect a significant influence of hallucinations on the relation extraction model's performance?} We address this question by evaluating the performance of models trained on datasets with varying levels of hallucinations, aiming to understand the presence and impact of different hallucination types. 
\textbf{RQ2: Can hallucinations be reliably detected?} %
To address this question, we 
develop and evaluate approaches for hallucination detection. 

Our findings reveal substantial declines in dataset quality and model performance due to hallucinations, with recall decreases ranging from 19.1\% to 39.2\%. This indicates that hallucinations notably compromise the ability of models to extract relations from texts. 
In this context, it is crucial to differentiate between relevant and irrelevant hallucinations. The former significantly affects performance, while the latter has a minimal impact. Furthermore, we develop two methods for identifying and eliminating hallucinations, achieving F1-scores of 83.8\% and 92.2\%. These methods not only remove hallucinations but also assist in estimating their prevalence. 

Overall, our contributions in this paper are as follows:\footnote{Our source code is available at \url{https://github.com/BigPanda042/Relation-Extraction-Hallucination-Study}.}

\begin{itemize}
    \item 
    \textbf{Analyzing the Impact of Hallucinations on Model Performance:} 
    We determine the effect of hallucinations on relation extraction models by training them on datasets with different levels of hallucinations and analyzing the performance discrepancies observed.
    \item 
    \textbf{Classifying Hallucinations:} 
    We categorize hallucinations into relevant and irrelevant types, and examine their impacts on datasets.
    \item 
    \textbf{Detecting Hallucinations:} 
    We evaluate language model-based methods for automatically detecting hallucinations. 
\end{itemize}

\section{Related Work}
\label{sec:Related-Work}

In this section, we first look at related work on creating synthetic training data. 
In the second part, we look at noisy data and hallucinations and how to recognize them.

\textbf{Generating Synthetic Data. }
\label{RW: Synthetic data generation}
Several data augmentation approaches have been proposed. Feng et al.~\cite{feng_survey_2021} differentiate between three main types:
(1) \textbf{Rule-based approaches} use algorithms to modify existing real-world datasets. Techniques such as synonym replacement, random insertion, swapping and deletion are used to significantly increase the volume of training data \cite{li_robust_2017, wei_eda_2019, wei_text_2021}.
(2) \textbf{Sample interpolation}, also known as \textbf{Mixed Sample Data Augmentation}, \cite{zhang_mixup_2018} interpolates data points to create more diverse and robust datasets for training language models \cite{yun_cutmix_2019, verma_manifold_2019, guo_nonlinear_2020, beckham_adversarial_2019, guo_sequence-level_2020}. 
Both approaches are limited by the fact that they are based on existing datasets. As a result, they are not able to introduce completely new features or vary the data types significantly, such as the relation types for relation extraction tasks. This can lead to the persistence of existing biases in the original datasets \cite{feng_survey_2021}.
(3) \textbf{Model-based approaches}, referred to as \textbf{Generative Data Augmentation (GDA)}, overcome these limitations. They are able to generate completely new and specific data points, independent of existing datasets. For example, the Control Prefixes model \cite{clive_control_2022} is characterized by the generation of text data from structured knowledge graphs using the WebNLG dataset \cite{webnlg_papersdt_2024, gardent_webnlg_2017}. Other notable implementations include the use of pretrained language models (PLMs), such as GPT-3.5, which have been successfully used 
to improve performance on relation extraction tasks \cite{xu_how_2023, gao_self-guided_2023, ye_zerogen_2022, josifoski_exploiting_2023}.

Josifoski et al. \cite{josifoski_exploiting_2023} developed a large synthetic dataset named Wiki-cIE for closed information extraction, utilizing GPT-3.5 with prompt engineering. This dataset, containing 1.8 million data points, serves as a robust alternative to both distantly supervised and directly supervised datasets in terms of size and quality. It is positioned closely in scale to the largest distantly supervised dataset, REBEL \cite{huguet_cabot_rebel_2021}. Importantly, the Wiki-cIE dataset offers enhanced quality, especially in the distribution of relation types and the accuracy of text annotations. Josifoski et al. demonstrate that relation extraction models trained on Wiki-cIE significantly outperform those trained on REBEL, attributing this advantage to the superior quality of their synthetic dataset. However, they do not specify which particular attributes of the datasets contribute to these performance differences. A notable quality difference is in the accuracy of the text annotations, suggesting that this aspect may be a critical factor in the observed improvements in model performance.

\textbf{Detecting and Assessing Noisy Data. }
\label{RW: Noisy Data}
Corrupted or noisy data, characterized by issues such as incorrect labels, 
affects language model training \cite{thulasidasan_combating_2019, ma_normalized_2020, liu_peer_2020, filippova_controlled_2020, tan_revisiting_2022, stoica_re-tacred_2021}. Several strategies have been developed to address noisy data in datasets. Techniques include resampling \cite{yu_how_2019}, loss reweighting \cite{thulasidasan_combating_2019}, and label correction \cite{ma_dimensionality-driven_2018}. Additionally, some approaches advocate training models using noise-robust loss functions \cite{ma_normalized_2020,liu_peer_2020}, with a notable recent development being a noise-robust re-weighting framework \cite{gao_self-guided_2023}. While these methods effectively mitigate the impact of noisy data or reduce its presence, they do not specifically explore the influence of hallucinations within synthetic training data on the performance of relation extraction models.

\textbf{Analyzing Hallucinations. }
\label{RW: Hallucinations}
Ji et al. \cite{ji_survey_2023} provide an overview on hallucinations, including relevant data-to-text use cases.  
The authors 
distinguish between two types of hallucinations: intrinsic and extrinsic. \textbf{Intrinsic hallucinations} are false information in texts that contradict the annotations, while \textbf{extrinsic hallucinations} consist of additional information in the texts that is not supported by the annotations.
There exist several approaches to detect both types of errors. Typical textual similarity metrics such as BLEU or ROUGE are unsuitable for the detection of hallucinations \cite{reiter_structured_2018, falke_ranking_2019}.
Other approaches can be divided into statistical and model-based methods. Statistical approaches \cite{wang_towards_2020, shuster_retrieval_2021, martindale_identifying_2019} focus primarily on lexical information, i.e., the specific words used, and therefore cannot adequately take syntactic or semantic variations into account. Thus, the more relevant alternatives are model-based approaches. Liu et al. \cite{liu_towards_2021} use named entity recognition to extract the entities from a text and compare them with those in the annotated table. The number of hallucinations is then based on the difference between annotated and found entities. Dušek and Kasner \cite{dusek_evaluating_2020} have developed an approach that uses a natural language based inference method. It compares the input data and the output text in both directions and can thus detect omissions or hallucinations. The last methods to be mentioned are the language model-based approaches by Filipova \cite{filippova_controlled_2020} and Tian et al. \cite{tian_sticking_2020}. However, these methods provide results that either focus on table-to-text generation or are not precise enough for our needs. While the presented methods contribute to the task of detecting hallucinations, none of them examines the exact influence of hallucinations on training performance or attempts to differentiate between different types of 
hallucinations. 

\section{Evaluation}
\label{sec:approach}

The concept of \textbf{hallucinations} lacks a universally accepted definition \cite{ji_survey_2023, filippova_controlled_2020, nie_simple_2019, ye_cognitive_2023, shuster_retrieval_2021}. 
Figure \ref{fig:example_hallucination} provides an example of a hallucination. In this scenario, a GDA model, tasked with generating text from the input triple ('Ted', 'lives in', 'New York'), should ideally produce 'Ted lives in the city of New York.' Instead, the model might extend this to 'Ted lives in the city of New York, which has a population of 8.4 million inhabitants.' This addition introduces an unsupported triple ('New York', 'has', '8.4 million inhabitants'), which is a hallucination.

Formally, we define hallucinations $H$ as the set difference $ H = TC \setminus T$ between the set of triples $T$ and the triples $TC$ that are actually generated in the text $S$.

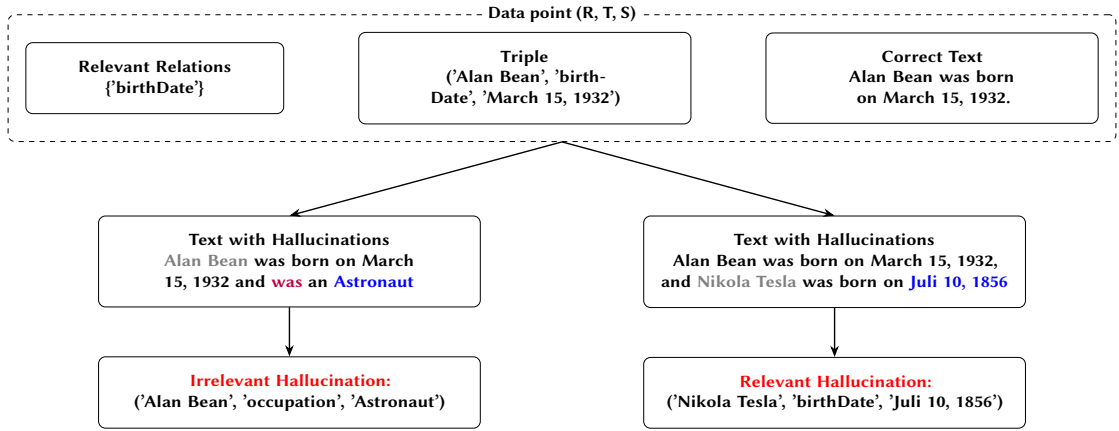
\begin{figure}[tb]
    \centering
\noindent
\resizebox{\textwidth}{!}{
\begin{tikzpicture}[
    box/.style={rectangle, draw=black, rounded corners, minimum height=1cm, minimum width=4cm, align=center, inner sep=10pt, font=\bfseries},
    dashedbox/.style={rectangle, draw=black, dashed, rounded corners, inner sep=10pt, fit=#1},
    arrow/.style={-Stealth, thick},
    textemph/.style={text=red, font=\bfseries},
    textnormal/.style={text=black, font=\normalfont}
]

\node[box, text width=4.5cm] (relevant) {
    \textbf{Relevant Relations} \\ \{'birthDate'\}
};

\node[box, right=1.5cm of relevant, text width=6cm] (triple) {
    \textbf{Triple} \\ ('Alan Bean', 'birthDate', 'March 15, 1932')
};

\node[box, right=1.5cm of triple, text width=6cm] (correcttext) {
    \textbf{Correct Text} \\ Alan Bean was born on March 15, 1932.
};

\node[dashedbox={(relevant) (triple) (correcttext)}] (datapointbox) {};

\node[below=-0.3cm of datapointbox.north, fill=white, rounded corners] {\textbf{Data point (R, T, S)}};

\node[box, below=1.5cm of datapointbox, text width=7cm, xshift=-5.5cm] (hallucination1) {
    \textbf{Text with Hallucinations} \\ 
    \textcolor{gray}{Alan Bean} was born on March 15, 1932 and \textcolor{purple}{was} an \textcolor{blue}{Astronaut}
};

\node[box, below=1.5cm of datapointbox, text width=7cm, xshift=5.5cm] (hallucination2) {
    \textbf{Text with Hallucinations} \\ 
    Alan Bean was born on March 15, 1932, and \textcolor{gray}{Nikola Tesla} was born on \textcolor{blue}{Juli 10, 1856}
};

\node[box, below=1cm of hallucination1, text width=7cm] (irrelevant) {
    \textcolor{red}{\textbf{Irrelevant Hallucination:}} \\ ('Alan Bean', 'occupation', 'Astronaut')
};

\node[box, below=1cm of hallucination2, text width=7cm] (relevanthallucination) {
    \textcolor{red}{\textbf{Relevant Hallucination:}} \\ ('Nikola Tesla', 'birthDate', 'Juli 10, 1856')
};

\draw[arrow] (datapointbox.south) -- (hallucination1.north);
\draw[arrow] (datapointbox.south) -- (hallucination2.north);

\draw[arrow] (hallucination1.south) -- ++(0,-0.75) -- (irrelevant.north);
\draw[arrow] (hallucination2.south) -- ++(0,-0.75) -- (relevanthallucination.north);

\end{tikzpicture}
}
    \caption{Examples of hallucinations, showing the distinction between relevant hallucinations (e.g., relation type 'birthDate') and irrelevant hallucinations (e.g., relation type 'occupation').}
    \label{fig:relevant_irrelevant_hallucinations}
\end{figure}

We differentiate between \textbf{relevant and irrelevant hallucinations} \cite{clive_control_2022, yan_partition_2021} in relation extraction models, as illustrated in Figure \ref{fig:relevant_irrelevant_hallucinations}. Relevant hallucinations occur when the text expresses triples with relation types that are relevant (i.e., included in the schema) but absent from the annotations. For example, if a model is trained exclusively to detect birth dates in texts, only triples related to birth dates are considered relevant. Conversely, irrelevant hallucinations involve relations that the model is designed to ignore, as they do not pertain to its trained focus.

In the following, we first analyze the influence of hallucinations on synthetic training datasets for relation extraction. 
We then consider the automatic detection 
of hallucinations in relation extraction datasets. %

\subsection{Evaluating the Effects of Hallucinations}
\label{sec:eval-effects-hallucinations} %

\subsubsection{Influence of Relevant Hallucinations on Document Level}
\label{Eval: Influence of Relevant Hallucinations on Document-level}

\begin{figure}[tb]
    \centering
\noindent
\resizebox{0.65\textwidth}{!}{
\begin{tikzpicture}[
    box/.style={rectangle, draw=black, rounded corners, minimum height=1cm, minimum width=4cm, align=center, text width=3cm, inner sep=8pt, font=\bfseries},
    arrow/.style={-Stealth, thick},
    oval/.style={ellipse, draw=black, minimum height=1cm, minimum width=3cm, align=center, font=\bfseries},
    ]

\node[box, text width=5cm] (datasetA) {
    \textbf{Dataset \textcolor{blue}{A}} \\
    Re-DocRED \\
    Fewer Hallucinations
};

\node[box, below=2cm of datasetA, text width=5cm] (datasetB) {
    \textbf{Dataset \textcolor{red}{B}} \\
    DocRED \\
    More Hallucinations
};

\node[box, right=2cm of datasetA, text width=5cm] (modelA) {
    \textbf{Model \textcolor{blue}{DREEAM$_A$}}
};

\node[box, right=2cm of datasetB, text width=5cm] (modelB) {
    \textbf{Model \textcolor{red}{DREEAM$_B$}}
};

\node[oval, below=1cm of modelA, text width=2cm] (comparison) {
    \textbf{Comparison}
};

\draw[arrow] (datasetA.east) -- node[above] {training} (modelA.west);
\draw[arrow] (datasetB.east) -- node[above] {training} (modelB.west);

\draw[arrow] (modelA.south) -- ++(0,-0.5) -- (comparison.north);

\draw[arrow] (modelB.north) -- ++(0,0.5) -- (comparison.south);

\end{tikzpicture}
}
    \caption{Illustration of the approach used for testing the influence of hallucinations at document-level with the datasets A and B and the language model approach DREEAM.}
    \label{fig:usedDataSets_for_evaluation}
\end{figure}
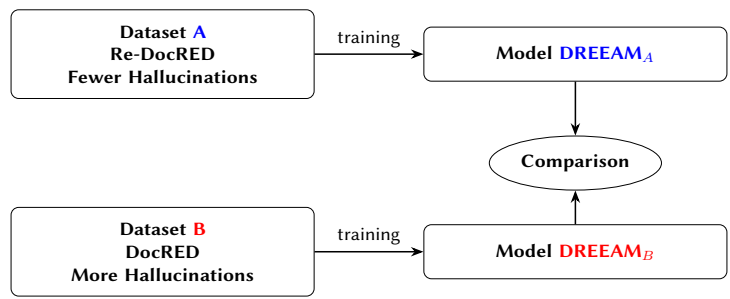

\newcolumntype{C}{>{\centering\arraybackslash}X}
\begin{table}[tb]
    \centering
    \caption{Key characteristics of Re-DocRED and DocRED.}
    \label{tab:difference_DocRed_Re-DocRed}
    \begin{small}
    \begin{tabular}{l| rrr| r r}
        \toprule
        \multirow{2}{*}{ } & \multicolumn{3}{c|}{\textbf{Re-DocRED}} & \multicolumn{2}{c}{\textbf{DocRED}} \\ 
          &  Train & Dev & Test & Train & Dev \\ \midrule
             Documents  & 3,053 & 500 & 500 & 3,053 & 1,000 \\
             Avg. Triples & 28.1 & 34.6 & 34.9 & 12.5 & 12.3 \\
             Avg. Sentences & 7.9 & 8.2 & 7.9 & 7.9 & 8.1 \\
        \bottomrule
    \end{tabular}
    \end{small}
\end{table}

In this subsection, we focus on evaluating the impact of \textit{relevant} hallucinations on \textit{document-level} relation extraction. 

\newpage 
\textbf{Datasets.} As presented in Figure~\ref{fig:usedDataSets_for_evaluation}, 
we select two datasets: 

\begin{enumerate}
 \item \textbf{Dataset A} %
is characterized by fewer hallucinations. Specifically, we employ the Re-DocRED dataset~\cite{tan_revisiting_2022}, as is known for its extensive use and minimal irrelevant content. %
 \item \textbf{Dataset B} %
contains a significant presence of relevant hallucinations. We use the DocRED dataset~\cite{yao_docred_2019}, an earlier version of the Re-DocRED dataset known for its incomplete annotations and the consequent prevalence of relevant hallucinations.
\end{enumerate}
The differences between 
dataset $A$, Re-DocRED, and $B$, DocRED, 
are outlined in Table~\ref{tab:difference_DocRed_Re-DocRed}. %

\textbf{Relation Extraction Model.} We select the DREEAM model~\cite{ma_dreeam_2023}, which has achieved top performance on the DocRED and Re-DocRED datasets~\cite{docred_papers_2024,redocred_papers_2024}.  
This model is 
optimized for compatibility with both datasets, thereby obviating the need for further modifications.

The model is initially trained on Dataset $A$ and $B$ to produce two tailored versions: DREEAM$_{A}$ and DREEAM$_{B}$. These models are then evaluated on the respective test portions of the datasets and benchmarked against the findings of Ma et al.~\cite{ma_dreeam_2023}. Although the standard practice for DocRED involves using a development dataset for parameter tuning and testing, we adopt it as our test dataset. The ultimate comparison of DREEAM$_{A}$ and DREEAM$_{B}$'s performance is conducted using the same test dataset A$^{test}$, which is 
free of hallucinations 
\cite{tan_revisiting_2022}.

\begin{table}[tb]
\centering 
        \caption{Precision, recall, and F1-score of the DREEAM model trained on A and B and evaluated on A$^{test}$ or B$^{test}$. Each value is the average of five configurations.}
    \label{fig:DREEAM_results}
    \begin{small}
    \begin{tabular}{ ll | rrr}
     \toprule
     Model& Test Dataset &Precision & Recall & F1\\
     \midrule
     DREEAM$_{A}$ & A$^{test}$ & 89.87 & 68.79 & 77.93 \\
     DREEAM$_{B}$  & A$^{test}$ & 94.86 & 29.59 & 45.11 \\
     \midrule
     DREEAM$_{B}$  & B$^{test}$ & 67.51 & 59.79 & 63.42 \\
     \bottomrule
    \end{tabular}
    \end{small}
\end{table}

\textbf{Evaluation Results and Discussion:~}
Table \ref{fig:DREEAM_results} shows the evaluation results, revealing a significant discrepancy between the two model configurations. Notably, the recall differs strongly, which is also reflected in the F1-score. In the case of relation extraction, the recall measures the ratio of correctly extracted triples compared to all relevant triples that should have been extracted. Since DREEAM$_{B}$ was trained on data where the triples in the annotation do not accurately reflect the text's triples, it learned that not all triples must be extracted to obtain a correct solution. This results in a lower recall, 
as expected. 

The precision, however, surprisingly increases for DREEAM$_{B}$ when evaluated on A$^{test}$, compared to the evaluation on B$^{test}$, to an even higher value than for DREEAM$_{A}$. 
This difference 
is most likely due to the wrong test dataset. The model most likely extracted true positives, but since the test dataset is incorrect, those correct triples were not present in the test annotation and counted as false positives. Those false positives became true positives through the correct A$^{test}$, and the precision increased. Nevertheless, this cannot explain why the precision increased further than the precision of DREEAM$_{A}$.
One potential reason 
is that DREEAM$_{B}$ tends to extract fewer triples than DREEAM$_{A}$. DREEAM$_{B}$ was trained on a dataset with generally fewer triples in T but the same texts and thus learned to extract fewer triples. 
Another possibility presents the relation type distribution. 
In $A$, the number of underrepresented relation types may have increased, or new, more difficult ones to extract correctly may have been introduced.

\subsubsection{Influence of Relevant Hallucinations on Sentence Level}
\label{Eval: Influence of Relevant Hallucinations on Sentence-Level}

\textbf{Datasets.} 
We now require datasets on the sentence-level. 
We use the WebNLG \cite{gardent_webnlg_2017} dataset for $A$, 
a widely used knowledge graph-to-text dataset \cite{webnlg_papersdt_2024, clive_control_2022, wang_stage-wise_2021, aghajanyan_htlm_2021}. 
Based on 
an own analysis and to the best of our knowledge, the dataset is free of relevant hallucinations \cite{zeng_extracting_2018}. 

The first variant for $B$, $B_{MT}$, is created to ensure direct comparability to the document-level datasets used above and to prevent biases by controlling the creation process. 
To accomplish this, we delete one triple from each of $A$'s data points, given that at least two triples are present. 
We randomly select which triple to delete to avoid any bias regarding the position of missing information. This ensures that the same texts are kept in both A and B$_{MT}$ but with different annotations. In total, we delete 28.1\% of all triples, corresponding to a 39.0\% hallucination rate
(calculated by dividing the total number of triples in all the texts by the total number of triples in all the annotations).  

Additionally, we create B$_{LT}$ to ensure that measured differences cannot solely be attributed to B$_{MT}$ just having fewer triples in the annotation. We add the text (of an unrelated data point) that contains no identical triples in the annotation to each data point of B$_{LT}$. This way, we can include relevant information in the text without altering the annotations. 

\textbf{Relation Extraction Model. }
We use the state-of-the-art PFN model \cite{yan_partition_2021,webnlg_papersre_2024}. 
Our initial step involved a preliminary experiment similar to the one described in Section 3.1.1. 
We adapted PFN to dataset A, resulting in PFN$_{A}$, and assessed its performance on the A$^{test}$ dataset. The F1-score differences were minor, averaging less than 0.6\% in variation, which we deemed acceptable given the unknown variance in Yan et al.'s results. Subsequently, we fine-tuned PFN on datasets B$_{MT}$ and B$_{LT}$, producing PFN$_{B_{MT}}$ and PFN$_{B{LT}}$, respectively. Both models were then evaluated against the original test dataset.  

\begin{table}[tb]
\centering 
        \caption{Precision, recall and F1-score of PFN models trained on datasets with relevant hallucinations $B_{MT}$ and $B_{MT}$ compared to the original dataset A. Each value is the average of five trained configurations evaluated on $A^{test}$.}
        \begin{small}
    \begin{tabular}{ l | rrr }
     \toprule
     Model& Precision & Recall & F1\\
     \midrule
     PFN$_{A}$  & 93.07 & 92.97 & 93.02 \\
     PFN$_{B_{MT}}$ & 91.41 & 73.87 & 81.70 \\
     PFN$_{B_{LT}}$ & 93.12 & 72.95 & 81.40 \\
     \bottomrule
    \end{tabular}
    \end{small}
    \label{fig:PFN_results_relevant}
\end{table}

\begin{table}[tbp]
    \centering
      \caption{Differences between models trained on the datasets A and B on sentence-level (B$_{MT}$ results used for B) and document-level.}
      \begin{small}
       \renewcommand{\arraystretch}{1.2}
    \begin{tabular}{l r r r r}
        \toprule
        \textbf{Relation Extraction Task} & \textbf{Dataset} & \textbf{Precision} & \textbf{Recall} & \textbf{F1} \\ \midrule
        \multirow{2}{*}{Sentence-Level} & A & 93.07 & 92.97 & 93.02  \\ 
         & B & 91.41 & 73.87 & 81.70 \\ \cline{2-5}
         & $\Delta$ & 1.66 & 19.10 & 11.32 \\ \midrule
        \multirow{2}{*}{Document-Level} & A & 89.87 & 68.79 & 77.93 \\ 
         & B$_{MT}$ & 94.86 & 29.59 & 45.11 \\ \cline{2-5}
         & $\Delta$ & 5.01 & 39.20 & 32.82 \\ \bottomrule
    \end{tabular}
    \end{small}
    \label{tab:difference_SL_vs_DL}
\end{table}

\textbf{Evaluation Results and Discussion. }
Table \ref{fig:PFN_results_relevant} reveals performance differences between PFN$_{A}$ and PFN$_{B_{MT}}$. Specifically, recall diminishes by an average of 19.1\%, while precision declines by 2.29\%, a decrease deemed statistically significant through a paired t-test at a 95\% confidence level.
Regarding PFN$_{B_{LT}}$, recall is similarly reduced by 19.98\%. Conversely, there is a marginal increase in precision of 0.06\%. 
Consequently, the F1-scores for PFN$_{B_{MT}}$ and PFN$_{B_{LT}}$ decrease by 11.32\% and 11.62\%, respectively.

These findings underline that the persistence of triples in longer texts within B$_{LT}$ does not counterbalance the reduced training data volume. As discussed in Section 3.1.1, 
only the variation in hallucination rates across datasets explains the altered recall rates. 

Differences remain substantial in recall between document-level and sentence-level extraction, as summarized in Table \ref{tab:difference_SL_vs_DL}. Document-level recall decreases nearly twice as much in absolute terms and three times in relative terms compared to sentence-level, primarily due to differing hallucination rates. Table \ref{tab:difference_DocRed_Re-DocRed} shows that Re-DocRED contains three times more triples per annotation than DocRED, a stronger contrast than observed between A and B$_{MT}$. Yet, without control over document-level dataset creation, a definitive causality cannot be verified here. %

Contrary to expectations, precision varies significantly across the experiments. Notably, a 5\% difference at the document-level, as indicated in Table \ref{fig:DREEAM_results}, diverges from the sentence-level findings between PFN$_{A}$ and PFN$_{B_{MT}}$. This discrepancy suggests potential document-specific effects or dataset variances not previously accounted for. Given the controlled modifications in B$_{MT}$, these results are considered more reliable, highlighting a distinct decrease in precision between PFN$_{A}$ and PFN$_{B_{MT}}$.

\subsubsection{Differences Between Relevant and Irrelevant Hallucinations}
\label{Eval: Differences Between Relevant and Irrelevant Hallucinations}

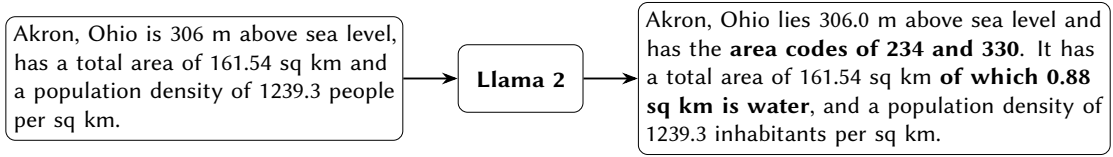
\begin{figure}[tbp]
    \centering
\noindent
\resizebox{\textwidth}{!}{
\begin{tikzpicture}[
    box/.style={rectangle, draw=black, rounded corners, minimum height=1cm, minimum width=4.5cm, align=left, text width=4.5cm},
    emph/.style={text=red, font=\bfseries},  %
    arrow/.style={-Stealth, thick}
]

\node[box, text width=5.6cm] (left) { %
    Akron, Ohio is 306 m above sea level, has a total area of 161.54 sq km and a population density of 1239.3 people per sq km.
};

\node[rectangle, draw=black, rounded corners, minimum height=1cm, text width=1.6cm, align=center, right=0.8cm of left] (middle) {
    \textbf{Llama 2}
};

\node[box, right=0.8cm of middle, text width=6.7cm] (right) { %
    Akron, Ohio lies 306.0 m above sea level and has the \textbf{area codes of 234 and 330}. 
    It has a total area of 161.54 sq km \textbf{of which 0.88 sq km is water}, and a population density of 1239.3 inhabitants per sq km.
};

\draw[arrow] (left) -- (middle);
\draw[arrow] (middle) -- (right);

\end{tikzpicture}
}
   \caption{Illustration of how LLAMA2-13b-chat may introduce hallucinations into a text.}
    \label{fig:LLama_demonstration_example}
\end{figure}

In this subsection, we evaluate the influence of relevant and irrelevant hallucinations on the sentence level.

\textbf{Dataset.} 
We keep $A$ the same as it can serve as the dataset with fewer hallucinations. On the other hand, $B$ needs to contain \textit{irrelevant} hallucinations instead of \textit{relevant} ones. We also create another test dataset for testing whether the newly trained models only extract the first part of a text and ignore the rest. 
To that end, we use the chat version of LLAMA2 \cite{touvron_llama_2023} to add irrelevant hallucinations to each of $A$'s data points. The LLM takes text as input and returns the same text but with additional information. 
To create additional dataset variants, we adjust the prompt by adding or removing specific instructions. This allows us to partly control the amount and type of information added. In total, we produce five modified WebNLG datasets, with the only difference being the prompt we use for the creation process.

\textbf{New Test Dataset. }
We modify the test dataset to assess if text length affects model performance. Each data point in A$^{test}$ is altered by fusing two data points (i.e., concatenating S and merging T), creating a test set with longer texts and more triples per data point without adding new hallucinations.

\begin{table}[tb]
    \caption{Precision, recall, and F1-score of PFN models trained on datasets with irrelevant hallucinations (B) compared to the original dataset (A), both evaluated on the original A$^{test}$ dataset. B consists of the average and standard deviation of five different datasets, with five model configurations trained on each of them.}
    \begin{small}
    \begin{tabular}{ l | lll }
     \toprule
     Model& Precision & Recall & F1\\
     \midrule
     PFN$_{A}$  & 93.07 & 92.97 & 93.02 \\
     PFN$_{B_{fused}}$ & 92.55 $\pm 0.26$ & 90.99 $\pm 1.15$ & 91.76 $\pm 0.65$ \\
     \bottomrule
    \end{tabular}
    \end{small}
    \label{fig:PFN_results_irrelevant}
\end{table}

\textbf{Language Model.} 
We fine-tune PFN on each of the five modified WebNLG datasets. %

\textbf{Evaluation Results on Original Test Dataset. } 
The results are presented in Table \ref{fig:PFN_results_irrelevant}. 
Since all the results of the modified WebNLG datasets are similar to each other, 
we present with B$_{fused}$ 
an average of all five 
(all results are in our repository). 
The evaluation on the original A$^{test}$ dataset shows that the recall drops for an average of 1.98\% and the F1-score for 1.26\% (statistically significant differences using the paired t-test and a confidence interval of 0.95).

\begin{table}[tb]
  \centering 
      \caption{Precision, recall, and F1-score of PFN 
      trained on A and B, both evaluated on the new test dataset.} 
    \label{fig:PFN_results_irrelevant_newTest}
    \begin{small}
    \begin{tabular}{ l | rrr }
     \toprule
     Model& Precision & Recall & F1\\
     \midrule
     PFN$_{A}$  & 65.14 & 88.49 & 75.02 \\
     PFN$_{B}$ & 63.74 & 86.99 & 73.56 \\
     \bottomrule
    \end{tabular}
    \end{small}
\end{table}

\textbf{Evaluation Results on New Test Dataset. }
The results are presented in Table \ref{fig:PFN_results_irrelevant_newTest}. They indicate a similar difference in the performance of PFN$_{A}$ and PFN$_{B}$ between the evaluation on the altered and the original test dataset. The recall difference is not statistically significant while the f1-score and precision are significant (using a paired t-test and confidence interval of 0.95).

\textbf{Discussion. }
The presented findings indicate that whether we extensively increase the information content, keep it a bit shorter, or create more similar information, the different prompts and irrelevant hallucinations have only a minor impact on the trained relation extraction models. Through the evaluation on the new test dataset, we observed that the small differences between PFN$_{A}$ and PFN$_{B_{fused}}$ cannot be attributed to the fact that PFN$_{B_{fused}}$ learned to ignore the back part of the natural language texts (which contains the newly added hallucinations) and only extracts triples from the front part. This is evident in the statistically insignificant difference between the recall of PFN$_{A}$ and PFN$_{B}$ evaluated on the new test dataset. The significant results in precision and F1-score are not further relevant for us. 

Overall, there is a minor impact of irrelevant hallucinations in relation extraction models because these models are trained to prioritize and extract only those relations classified under relevant relation types, effectively disregarding all others categorized as irrelevant relation types. Thus, 
irrelevant hallucinations 
are systematically ignored during the training process. 

The observed differences in model performance, despite expectations, may stem from two potential factors that require further investigation. The first possibility is that Llama 2 occasionally introduces relevant relations in what is mostly uncontrolled information. 
The second possibility is an increase in errors by the relation extraction model due to processing a larger volume of text, regardless of its relevance. 
Further experiments are needed in this regard. 

Based on these results and explanations, we can confirm the assumption that relevant hallucinations in (synthetic) training data have a much stronger impact on the performance of relation extraction models trained on them than irrelevant hallucinations. Therefore, irrelevant hallucinations can mostly be neglected regarding the influence on training performance of relation extraction models. This also means that when creating datasets or improving annotation quality, removing relevant hallucinations should be the priority.

\subsection{Evaluating Hallucination Detection} %
\label{Eval: Evaluating Metrics for Hallucination Detection and Quantification} %

We consider two approaches of 
hallucination detection, 
as outlined in the following.  %

\subsubsection{Named Entity Recognition-Based Hallucination Detection}
\label{Eval: Evaluating the NER Metric for Hallucination Detection}

A first approach for hallucination detection was suggested by Liu et al. \cite{liu_towards_2021} and involves named entity recognition (NER). This approach extracts entities from a text and compares them to the entities in the corresponding triples. Entities found in the text but absent from the triples are identified as hallucinations.

\textbf{Dataset:} %
The sentence-level WebNLG dataset version v3.0 \cite{gardent_creating_2017} serves as the basis for this work. 
The dataset includes annotations with one to seven triples. %

\textbf{Model:}  %
We decided to use the widely used SpaCy model 
given its wide usage and solid performance. 
Through preliminary tests, we can confirm the theory 
that the entities extracted by the model from S are often correct but not equivalent to those in the annotation T. This can result in cases where, for example, 'Alan Bean' is in T but only 'A. Bean' is extracted from S, which essentially means the same thing. To solve this problem, we use the sentence similarity model \textit{all-mpnet-base-v2}~\cite{noauthor_sentence-transformersall-mpnet-base-v2_2024} 
to compare the extracted and annotated entities.

\textbf{Evaluation Setup: }
We use the precision, recall, and F1-score for the evaluation. We classify 'hallucination-free' texts that are correctly accepted as true positives. 

For our experiments, we utilize 3,000 data points sampled from D. For each data point, we randomly select one correct text and one hallucination to maintain a balanced ratio between the two. The hallucination text can contain one to six hallucinated triples. Additionally, we conduct a hyperparameter sweep across all acceptance thresholds ranging from 0.05 to 0.95 (inclusive) in 0.05 increments to find the best-performing threshold for the sentence similarity model.

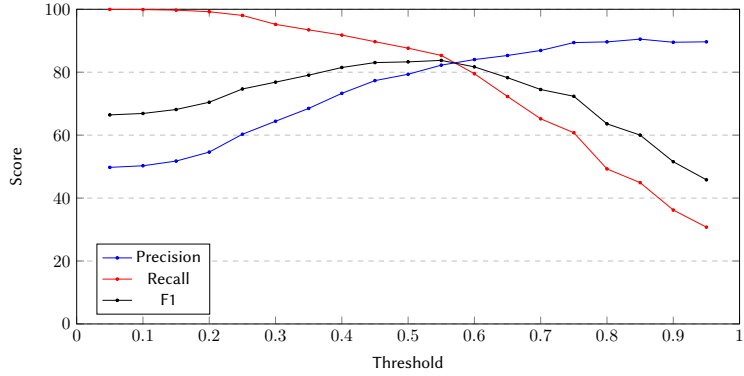
\begin{figure}[tbp]
\centering 
    \begin{tikzpicture}[scale = .65]
    \begin{axis}[
        width=15cm,
        height= 8cm,
        xlabel={Threshold},
        ylabel=Score,
        xmin=0, xmax= 1,
        ymin=0, ymax=100,
        xtick={0,0.1,0.2,0.3,0.4,0.5,0.6,0.7,0.8,0.9,1},
        ytick={0,20,40,60,80,100},
        legend pos=south west,
        legend entries={Precision,Recall,F1},
        ymajorgrids=true,
        grid style=dashed,
        ]
      \addplot[color = blue,  mark=10-pointed star, mark size=1pt] table [x=Threshold,y=Recall] {images/txts/SpaCy_alone.txt};
      \addplot[color = red, mark=10-pointed star, mark size=1pt] table [x=Threshold,y=Precision] {images/txts/SpaCy_alone.txt};
      \addplot[color = black, mark=10-pointed star, mark size=1pt] table [x=Threshold,y=F1] {images/txts/SpaCy_alone.txt};
    \end{axis}
    \end{tikzpicture}
    \caption{Precision, recall, and F1-score of the NER metric at varying acceptance thresholds. The highest F1-score is achieved at a threshold of 0.55.}
    \label{fig:Spacy_results}
\end{figure}

\textbf{Evaluation Results: }
Figure \ref{fig:Spacy_results} shows a climbing precision and falling recall with increasing threshold. Given those trends, the F1-score increases with a higher threshold up to 0.55. After this, it falls until the end.
At the peak of 0.55, the precision, recall, and F1-score are 85.34, 82.25, and 83.76\%, respectively. 
Given this, the overall results obtained from the tests seem satisfactory. Out of all the texts classified as 'clean,' around 85.34\% were correctly identified as clean. Similarly, among all the tested clean texts, 82.25\% were accurately classified as 'clean.' With that performance, the approach can be used to detect hallucinations and provide an approximate understanding of the amount of hallucinations in datasets. 

The presented approach has several limitations. One limitation is that the equivalence between extracted and annotated entities depends on the sentence similarity model, making it unclear how many entities were incorrectly accepted or rejected. A fixed threshold is also needed to define equivalence, with the best results found at 0.55, indicating significant differences between extracted entities and annotations. This complicates model evaluation, and the issue can vary across datasets due to differing annotation formats. A potential solution is to use textual entailment instead of sentence similarity to assess entity matches.

\subsubsection{Textual Entailment Approach for Hallucination Detection}
\label{eval: Evaluating the Textual Entailment Metric for Hallucination Detection} %

Another approach, inspired by Dušek and Kasner \cite{dusek_evaluating_2020}, uses an entailment model $M$ to check if a sentence $S$ contains the same information as a set of triples $T$. The triples $t \in T$ are combined into a single sentence using conjunctions. If $M$ classifies $T$ as not entailed, it indicates hallucinations; if classified as entailed, $S$ is considered hallucination-free.

\textbf{Dataset: }
Since we evaluate a new approach on the same task as in the previous Section \ref{Eval: Evaluating the NER Metric for Hallucination Detection}, we do not need to adjust the dataset and can continue to use $D$.

\textbf{Model: } %
For this task, we focus on the \textit{roberta-large-mnli} \cite{liu_roberta_2019} and \textit{deberta-v2-xlarge-mnli} \cite{he_deberta_2021} models. Both models perform well on SQuAD 1.1/2.0 and various GLUE benchmarks. 

\textbf{Classifier: }
An entailment model can be used to test whether sentence S2 is part of sentence S1 or if the content of S1 implies the content of S2. 
We design the model to test for any hallucinations in S compared to T, from each data point of a dataset. 

The initial step involves pre-processing the triples, typically formatted as 'Entity\_1 | relation | Entity\_2' or as a three-element list. Here, we receive the input in the former format and replace all occurrences of '\_' and ' | ' with spaces. 
Next, the goal is to create a sentence, ST, that encompasses all triples from T and accurately conveys T's informative value. This is done by combining the pre-processed triples $t \in T$ into a single sentence, linked by the conjunction 'and.' 
Finally, M verifies whether S is entailed in ST. If the result is 'entailed,' S is deemed correct; otherwise, 'neutral' or 'contradictory' results signal the presence of hallucinations.

\textbf{Evaluation Setup: }
We test $M$ on 4,000 sampled data points from $D$. Each sample comprises an annotation $T$ and two texts, $S1 \in h$ and $S2 \in c$, while $h,c \subset D$. 
That means that two tests have to be conducted for each data point, one for each text. This results in 
8,000 classifications.
The evaluation is otherwise similar to the NER approach from the previous section. 

\textbf{Evaluation Results: } %
The results are presented in Table~\ref{tab:Entailment_result}. Both tested models outperformed the SpaCy model by 7.03\% to 8.39\% in F1-score using a straightforward sentence creation procedure for ST. The deberta-v2-xlarge-mnli model outperforms the roberta-large-mnli model by 1.36\% in F1-score, with the most significant difference in recall, while precision increases only slightly. When comparing our best hallucination detection approach to the method used by Dušek and Kasner \cite{dusek_evaluating_2020}, our approach performs significantly better, achieving a 13.75\% higher F1-score. However, this comparison should be considered with caution, as their study used an older version of the WebNLG dataset with many incorrect annotations.

An F1-score of 92.15\% demonstrates that it is possible to reliably classify each data point in a dataset as either containing hallucinations or being hallucination-free. With high recall and precision, most clean texts are correctly identified as such, and the error rate for texts wrongly identified as hallucination-free is under 10\%. 
This performance allows for the effective detection (and removal) of hallucinations in datasets, thereby significantly improving annotation quality. 

Despite the superior results compared to Dušek and Kasner \cite{dusek_evaluating_2020}, who also relied on natural language inference, 
the comparison must be approached with caution due to their use of an outdated WebNLG version with incorrect annotations affecting their outcomes. Furthermore, this approach is limited to the detection of hallucinations and to sentence-level datasets, similar to the constraints discussed for the NER metric.

\begin{table}[tb]
    \centering
\caption{Comparison of NER-based and textual entailment approaches for detecting hallucinations in relation extraction tasks.}
    \label{tab:Entailment_result}
\begin{small}
    \begin{tabular}{ l | r r r }
        \toprule
        Approach  & Precision & Recall & F1 \\
        \midrule
        NER with \textit{SpaCy}    & 85.34 & 82.25 & 83.76 \\
        NLI with \textit{roberta-large-mnli}   & 88.29 & 93.45 & 90.79 \\
        NLI with \textit{deberta-v2-xlarge-mnli}   & \textbf{90.05} & \textbf{94.35} & \textbf{92.15} \\
        NLI with \textit{Dušek and Kasner} \cite{dusek_evaluating_2020}    & 77.20 & 79.60 & 78.40 \\
        \bottomrule
    \end{tabular}
    \end{small}

\end{table}

\section{Conclusion}
\label{Conclusion}

In this paper, we analyzed the impact of hallucinations in synthetic training data on relation extraction tasks. 
Our evaluation revealed significant performance declines with recall reductions between 19.1\% and 39.2\%. This indicates that hallucinations notably compromise the ability of models to accurately extract relations from texts. We identified a distinction between relevant and irrelevant hallucinations, noting that the former significantly impairs performance, while the latter has a minimal impact.
Additionally, we developed methods for the detection (and thus mitigation) of hallucinations to improve data quality and, thus, model performance. Our approaches, 
successfully classified texts as either 'hallucinated' or 'clean,' 
with 
notable F1-scores of 83.8\% and 92.2\%. 
In the future, we will analyze the impact of hallucinations in datasets for other NLP tasks, such as entity and event extraction.


\bibliography{references}

\begin{thebibliography}{58}
\expandafter\ifx\csname natexlab\endcsname\relax\def\natexlab#1{#1}\fi
\providecommand{\url}[1]{\texttt{#1}}
\providecommand{\href}[2]{#2}
\providecommand{\path}[1]{#1}
\providecommand{\DOIprefix}{doi:}
\providecommand{\ArXivprefix}{arXiv:}
\providecommand{\URLprefix}{URL: }
\providecommand{\Pubmedprefix}{pmid:}
\providecommand{\doi}[1]{\href{http://dx.doi.org/#1}{\path{#1}}}
\providecommand{\Pubmed}[1]{\href{pmid:#1}{\path{#1}}}
\providecommand{\bibinfo}[2]{#2}
\ifx\xfnm\relax \def\xfnm[#1]{\unskip,\space#1}\fi
\bibitem[{Hern{\'{a}}ndez{-}Garc{\'{\i}}a and K{\"{o}}nig(2018)}]{hernandez-garcia_data_2020}
\bibinfo{author}{A.~Hern{\'{a}}ndez{-}Garc{\'{\i}}a}, \bibinfo{author}{P.~K{\"{o}}nig},
\newblock \bibinfo{title}{Data augmentation instead of explicit regularization},
\newblock \bibinfo{journal}{CoRR} \bibinfo{volume}{abs/1806.03852} (\bibinfo{year}{2018}). \URLprefix \url{http://arxiv.org/abs/1806.03852}. \href{http://arxiv.org/abs/1806.03852}{{\tt arXiv:1806.03852}}.
\bibitem[{Gao et~al.(2023)Gao, Pi, Lin, Xu, Ye, Wu, Zhang, Liang, Li, and Kong}]{gao_self-guided_2023}
\bibinfo{author}{J.~Gao}, \bibinfo{author}{R.~Pi}, \bibinfo{author}{Y.~Lin}, \bibinfo{author}{H.~Xu}, \bibinfo{author}{J.~Ye}, \bibinfo{author}{Z.~Wu}, \bibinfo{author}{W.~Zhang}, \bibinfo{author}{X.~Liang}, \bibinfo{author}{Z.~Li}, \bibinfo{author}{L.~Kong},
\newblock \bibinfo{title}{Self-guided noise-free data generation for efficient zero-shot learning},
\newblock in: \bibinfo{booktitle}{Proceedings of the Eleventh International Conference on Learning Representations}, ICLR'23, \bibinfo{year}{2023}.
\bibitem[{Wei and Zou(2019)}]{wei_eda_2019}
\bibinfo{author}{J.~Wei}, \bibinfo{author}{K.~Zou},
\newblock \bibinfo{title}{{EDA}: {Easy} {Data} {Augmentation} {Techniques} for {Boosting} {Performance} on {Text} {Classification} {Tasks}},
\newblock in: \bibinfo{booktitle}{Proceedings of the 2019 {Conference} on {Empirical} {Methods} in {Natural} {Language} {Processing} and the 9th {International} {Joint} {Conference} on {Natural} {Language} {Processing}}, EMNLP-IJCNLP'19, \bibinfo{address}{Hong Kong, China}, \bibinfo{year}{2019}, pp. \bibinfo{pages}{6382--6388}. \URLprefix \url{https://aclanthology.org/D19-1670}. \DOIprefix\doi{10.18653/v1/D19-1670}.
\bibitem[{Anaby{-}Tavor et~al.(2020)Anaby{-}Tavor, Carmeli, Goldbraich, Kantor, Kour, Shlomov, Tepper, and Zwerdling}]{anaby-tavor_not_2019}
\bibinfo{author}{A.~Anaby{-}Tavor}, \bibinfo{author}{B.~Carmeli}, \bibinfo{author}{E.~Goldbraich}, \bibinfo{author}{A.~Kantor}, \bibinfo{author}{G.~Kour}, \bibinfo{author}{S.~Shlomov}, \bibinfo{author}{N.~Tepper}, \bibinfo{author}{N.~Zwerdling},
\newblock \bibinfo{title}{Do not have enough data? deep learning to the rescue!},
\newblock in: \bibinfo{booktitle}{Proceedings of the 34th {AAAI} Conference on Artificial Intelligence}, AAAI'20, \bibinfo{publisher}{{AAAI} Press}, \bibinfo{year}{2020}, pp. \bibinfo{pages}{7383--7390}. \URLprefix \url{https://doi.org/10.1609/aaai.v34i05.6233}. \DOIprefix\doi{10.1609/AAAI.V34I05.6233}.
\bibitem[{Cabot and Navigli(2021)}]{huguet_cabot_rebel_2021}
\bibinfo{author}{P.~H. Cabot}, \bibinfo{author}{R.~Navigli},
\newblock \bibinfo{title}{{REBEL:} relation extraction by end-to-end language generation},
\newblock in: \bibinfo{booktitle}{Findings of the 2021 Conference on Empirical Methods in Natural Language Processing}, EMNLP'21, \bibinfo{year}{2021}, pp. \bibinfo{pages}{2370--2381}. \URLprefix \url{https://doi.org/10.18653/v1/2021.findings-emnlp.204}.
\bibitem[{Josifoski et~al.(2023)Josifoski, Sakota, Peyrard, and West}]{josifoski_exploiting_2023}
\bibinfo{author}{M.~Josifoski}, \bibinfo{author}{M.~Sakota}, \bibinfo{author}{M.~Peyrard}, \bibinfo{author}{R.~West},
\newblock \bibinfo{title}{Exploiting {Asymmetry} for {Synthetic} {Training} {Data} {Generation}: {SynthIE} and the {Case} of {Information} {Extraction}},
\newblock in: \bibinfo{booktitle}{Proceedings of the 2023 {Conference} on {Empirical} {Methods} in {Natural} {Language} {Processing}}, EMNLP'23, \bibinfo{address}{Singapore}, \bibinfo{year}{2023}, pp. \bibinfo{pages}{1555--1574}. \URLprefix \url{https://aclanthology.org/2023.emnlp-main.96}. \DOIprefix\doi{10.18653/v1/2023.emnlp-main.96}.
\bibitem[{Ribeiro et~al.(2021)Ribeiro, Schmitt, Schütze, and Gurevych}]{ribeiro_investigating_2021}
\bibinfo{author}{L.~F.~R. Ribeiro}, \bibinfo{author}{M.~Schmitt}, \bibinfo{author}{H.~Schütze}, \bibinfo{author}{I.~Gurevych},
\newblock \bibinfo{title}{Investigating {Pretrained} {Language} {Models} for {Graph}-to-{Text} {Generation}},
\newblock in: \bibinfo{booktitle}{Proceedings of the 3rd {Workshop} on {Natural} {Language} {Processing} for {Conversational} {AI}}, NLP4ConvAI@ACL'21, \bibinfo{address}{Online}, \bibinfo{year}{2021}, pp. \bibinfo{pages}{211--227}. \URLprefix \url{https://aclanthology.org/2021.nlp4convai-1.20}. \DOIprefix\doi{10.18653/v1/2021.nlp4convai-1.20}.
\bibitem[{Wang et~al.(2021)Wang, Yavuz, Lin, Ji, and Rajani}]{wang_stage-wise_2021}
\bibinfo{author}{Q.~Wang}, \bibinfo{author}{S.~Yavuz}, \bibinfo{author}{X.~V. Lin}, \bibinfo{author}{H.~Ji}, \bibinfo{author}{N.~F. Rajani},
\newblock \bibinfo{title}{Stage-wise fine-tuning for graph-to-text generation},
\newblock in: \bibinfo{booktitle}{Proceedings of the {ACL-IJCNLP} 2021 Student Research Workshop, {ACL} 2021, Online, JUli 5-10, 2021}, \bibinfo{publisher}{Association for Computational Linguistics}, \bibinfo{year}{2021}, pp. \bibinfo{pages}{16--22}. \URLprefix \url{https://doi.org/10.18653/v1/2021.acl-srw.2}. \DOIprefix\doi{10.18653/V1/2021.ACL-SRW.2}.
\bibitem[{Chen et~al.(2020)Chen, Yang, and Yang}]{chen_mixtext_2020}
\bibinfo{author}{J.~Chen}, \bibinfo{author}{Z.~Yang}, \bibinfo{author}{D.~Yang},
\newblock \bibinfo{title}{{MixText}: {Linguistically}-{Informed} {Interpolation} of {Hidden} {Space} for {Semi}-{Supervised} {Text} {Classification}},
\newblock in: \bibinfo{booktitle}{Proceedings of the 58th {Annual} {Meeting} of the {Association} for {Computational} {Linguistics}}, ACL'20, \bibinfo{address}{Online}, \bibinfo{year}{2020}, pp. \bibinfo{pages}{2147--2157}. \URLprefix \url{https://aclanthology.org/2020.acl-main.194}. \DOIprefix\doi{10.18653/v1/2020.acl-main.194}.
\bibitem[{Thakur et~al.(2021)Thakur, Reimers, Daxenberger, and Gurevych}]{thakur_augmented_2021}
\bibinfo{author}{N.~Thakur}, \bibinfo{author}{N.~Reimers}, \bibinfo{author}{J.~Daxenberger}, \bibinfo{author}{I.~Gurevych},
\newblock \bibinfo{title}{Augmented {SBERT}: {Data} {Augmentation} {Method} for {Improving} {Bi}-{Encoders} for {Pairwise} {Sentence} {Scoring} {Tasks}},
\newblock in: \bibinfo{editor}{K.~Toutanova}, \bibinfo{editor}{A.~Rumshisky}, \bibinfo{editor}{L.~Zettlemoyer}, \bibinfo{editor}{D.~Hakkani-Tur}, \bibinfo{editor}{I.~Beltagy}, \bibinfo{editor}{S.~Bethard}, \bibinfo{editor}{R.~Cotterell}, \bibinfo{editor}{T.~Chakraborty}, \bibinfo{editor}{Y.~Zhou} (Eds.), \bibinfo{booktitle}{Proceedings of the 2021 {Conference} of the {North} {American} {Chapter} of the {Association} for {Computational} {Linguistics}: {Human} {Language} {Technologies}}, \bibinfo{publisher}{Association for Computational Linguistics}, \bibinfo{address}{Online}, \bibinfo{year}{2021}, pp. \bibinfo{pages}{296--310}. \URLprefix \url{https://aclanthology.org/2021.naacl-main.28}. \DOIprefix\doi{10.18653/v1/2021.naacl-main.28}.
\bibitem[{Yoo et~al.(2021)Yoo, Park, Kang, Lee, and Park}]{yoo_gpt3mix_2021}
\bibinfo{author}{K.~M. Yoo}, \bibinfo{author}{D.~Park}, \bibinfo{author}{J.~Kang}, \bibinfo{author}{S.-W. Lee}, \bibinfo{author}{W.~Park},
\newblock \bibinfo{title}{{GPT3Mix}: {Leveraging} {Large}-scale {Language} {Models} for {Text} {Augmentation}},
\newblock in: \bibinfo{booktitle}{Findings of the 2021 Conference on Empirical Methods in Natural Language Processing}, EMNLP'21, \bibinfo{address}{Punta Cana, Dominican Republic}, \bibinfo{year}{2021}, pp. \bibinfo{pages}{2225--2239}. \URLprefix \url{https://aclanthology.org/2021.findings-emnlp.192}. \DOIprefix\doi{10.18653/v1/2021.findings-emnlp.192}.
\bibitem[{Ye et~al.(2023)Ye, Liu, Zhang, Hua, and Jia}]{ye_cognitive_2023}
\bibinfo{author}{H.~Ye}, \bibinfo{author}{T.~Liu}, \bibinfo{author}{A.~Zhang}, \bibinfo{author}{W.~Hua}, \bibinfo{author}{W.~Jia}, \bibinfo{title}{Cognitive {Mirage}: {A} {Review} of {Hallucinations} in {Large} {Language} {Models}}, \bibinfo{year}{2023}. \URLprefix \url{http://arxiv.org/abs/2309.06794}. \DOIprefix\doi{10.48550/arXiv.2309.06794}.
\bibitem[{Ji et~al.(2023)Ji, Lee, Frieske, Yu, Su, Xu, Ishii, Bang, Chen, Chan, Dai, Madotto, and Fung}]{ji_survey_2023}
\bibinfo{author}{Z.~Ji}, \bibinfo{author}{N.~Lee}, \bibinfo{author}{R.~Frieske}, \bibinfo{author}{T.~Yu}, \bibinfo{author}{D.~Su}, \bibinfo{author}{Y.~Xu}, \bibinfo{author}{E.~Ishii}, \bibinfo{author}{Y.~Bang}, \bibinfo{author}{D.~Chen}, \bibinfo{author}{H.~S. Chan}, \bibinfo{author}{W.~Dai}, \bibinfo{author}{A.~Madotto}, \bibinfo{author}{P.~Fung},
\newblock \bibinfo{title}{Survey of {Hallucination} in {Natural} {Language} {Generation}},
\newblock \bibinfo{journal}{ACM Computing Surveys} \bibinfo{volume}{55} (\bibinfo{year}{2023}) \bibinfo{pages}{1--38}. \URLprefix \url{http://arxiv.org/abs/2202.03629}. \DOIprefix\doi{10.1145/3571730}, \bibinfo{note}{arXiv:2202.03629}.
\bibitem[{Varshney et~al.(2023)Varshney, Yao, Zhang, Chen, and Yu}]{varshney_stitch_2023}
\bibinfo{author}{N.~Varshney}, \bibinfo{author}{W.~Yao}, \bibinfo{author}{H.~Zhang}, \bibinfo{author}{J.~Chen}, \bibinfo{author}{D.~Yu},
\newblock \bibinfo{title}{{A Stitch in Time Saves Nine: Detecting and Mitigating Hallucinations of LLMs by Validating Low-Confidence Generation}},
\newblock \bibinfo{journal}{CoRR} \bibinfo{volume}{abs/2307.03987} (\bibinfo{year}{2023}). \URLprefix \url{https://doi.org/10.48550/arXiv.2307.03987}. \DOIprefix\doi{10.48550/ARXIV.2307.03987}.
\bibitem[{Feng et~al.(2021)Feng, Gangal, Wei, Chandar, Vosoughi, Mitamura, and Hovy}]{feng_survey_2021}
\bibinfo{author}{S.~Y. Feng}, \bibinfo{author}{V.~Gangal}, \bibinfo{author}{J.~Wei}, \bibinfo{author}{S.~Chandar}, \bibinfo{author}{S.~Vosoughi}, \bibinfo{author}{T.~Mitamura}, \bibinfo{author}{E.~Hovy},
\newblock \bibinfo{title}{A {Survey} of {Data} {Augmentation} {Approaches} for {NLP}},
\newblock in: \bibinfo{booktitle}{Findings of the {Association} for {Computational} {Linguistics}}, {ACL}-{IJCNLP}'21, \bibinfo{address}{Virtual Event}, \bibinfo{year}{2021}, pp. \bibinfo{pages}{968--988}. \URLprefix \url{https://aclanthology.org/2021.findings-acl.84}. \DOIprefix\doi{10.18653/v1/2021.findings-acl.84}.
\bibitem[{Li et~al.(2017)Li, Cohn, and Baldwin}]{li_robust_2017}
\bibinfo{author}{Y.~Li}, \bibinfo{author}{T.~Cohn}, \bibinfo{author}{T.~Baldwin},
\newblock \bibinfo{title}{Robust {Training} under {Linguistic} {Adversity}},
\newblock in: \bibinfo{booktitle}{Proceedings of the 15th {Conference} of the {European} {Chapter} of the {Association} for {Computational} {Linguistics}}, EACL'17, \bibinfo{address}{Valencia, Spain}, \bibinfo{year}{2017}, pp. \bibinfo{pages}{21--27}. \URLprefix \url{https://aclanthology.org/E17-2004}.
\bibitem[{Wei et~al.(2021)Wei, Huang, Xu, and Vosoughi}]{wei_text_2021}
\bibinfo{author}{J.~Wei}, \bibinfo{author}{C.~Huang}, \bibinfo{author}{S.~Xu}, \bibinfo{author}{S.~Vosoughi},
\newblock \bibinfo{title}{Text {Augmentation} in a {Multi}-{Task} {View}},
\newblock in: \bibinfo{booktitle}{Proceedings of the 16th {Conference} of the {European} {Chapter} of the {Association} for {Computational} {Linguistics}}, EACL'21, \bibinfo{address}{Virtual Event}, \bibinfo{year}{2021}, pp. \bibinfo{pages}{2888--2894}. \URLprefix \url{https://aclanthology.org/2021.eacl-main.252}. \DOIprefix\doi{10.18653/v1/2021.eacl-main.252}.
\bibitem[{Zhang et~al.(2018)Zhang, Cisse, Dauphin, and Lopez-Paz}]{zhang_mixup_2018}
\bibinfo{author}{H.~Zhang}, \bibinfo{author}{M.~Cisse}, \bibinfo{author}{Y.~N. Dauphin}, \bibinfo{author}{D.~Lopez-Paz}, \bibinfo{title}{mixup: {Beyond} {Empirical} {Risk} {Minimization}}, \bibinfo{year}{2018}. \URLprefix \url{http://arxiv.org/abs/1710.09412}. \DOIprefix\doi{10.48550/arXiv.1710.09412}, \bibinfo{note}{arXiv:1710.09412}.
\bibitem[{Yun et~al.(2019)Yun, Han, Chun, Oh, Yoo, and Choe}]{yun_cutmix_2019}
\bibinfo{author}{S.~Yun}, \bibinfo{author}{D.~Han}, \bibinfo{author}{S.~Chun}, \bibinfo{author}{S.~J. Oh}, \bibinfo{author}{Y.~Yoo}, \bibinfo{author}{J.~Choe},
\newblock \bibinfo{title}{Cutmix: Regularization strategy to train strong classifiers with localizable features},
\newblock in: \bibinfo{booktitle}{Proceedings of the 2019 {IEEE/CVF} International Conference on Computer Vision}, ICCV'19, \bibinfo{publisher}{{IEEE}}, \bibinfo{year}{2019}, pp. \bibinfo{pages}{6022--6031}. \URLprefix \url{https://doi.org/10.1109/ICCV.2019.00612}. \DOIprefix\doi{10.1109/ICCV.2019.00612}.
\bibitem[{Verma et~al.(2019)Verma, Lamb, Beckham, Najafi, Mitliagkas, Lopez-Paz, and Bengio}]{verma_manifold_2019}
\bibinfo{author}{V.~Verma}, \bibinfo{author}{A.~Lamb}, \bibinfo{author}{C.~Beckham}, \bibinfo{author}{A.~Najafi}, \bibinfo{author}{I.~Mitliagkas}, \bibinfo{author}{D.~Lopez-Paz}, \bibinfo{author}{Y.~Bengio},
\newblock \bibinfo{title}{Manifold {Mixup}: {Better} {Representations} by {Interpolating} {Hidden} {States}},
\newblock in: \bibinfo{booktitle}{Proceedings of the 36th {International} {Conference} on {Machine} {Learning}}, ICML'19, \bibinfo{year}{2019}, pp. \bibinfo{pages}{6438--6447}.
\bibitem[{Guo(2020)}]{guo_nonlinear_2020}
\bibinfo{author}{H.~Guo},
\newblock \bibinfo{title}{Nonlinear {Mixup}: {Out}-{Of}-{Manifold} {Data} {Augmentation} for {Text} {Classification}},
\newblock \bibinfo{journal}{Proceedings of the AAAI Conference on Artificial Intelligence} \bibinfo{volume}{34} (\bibinfo{year}{2020}) \bibinfo{pages}{4044--4051}. \DOIprefix\doi{10.1609/aaai.v34i04.5822}.
\bibitem[{Beckham et~al.(2019)Beckham, Honari, Verma, Lamb, Ghadiri, Hjelm, Bengio, and Pal}]{beckham_adversarial_2019}
\bibinfo{author}{C.~Beckham}, \bibinfo{author}{S.~Honari}, \bibinfo{author}{V.~Verma}, \bibinfo{author}{A.~M. Lamb}, \bibinfo{author}{F.~Ghadiri}, \bibinfo{author}{R.~D. Hjelm}, \bibinfo{author}{Y.~Bengio}, \bibinfo{author}{C.~Pal},
\newblock \bibinfo{title}{On {Adversarial} {Mixup} {Resynthesis}},
\newblock in: \bibinfo{booktitle}{Advances in {Neural} {Information} {Processing} {Systems}}, volume~\bibinfo{volume}{32} of \textit{\bibinfo{series}{NeurIPS'19}}, \bibinfo{year}{2019}. \URLprefix \url{https://papers.nips.cc/paper/2019/hash/f708f064faaf32a43e4d3c784e6af9ea-Abstract.html}.
\bibitem[{Guo et~al.(2020)Guo, Kim, and Rush}]{guo_sequence-level_2020}
\bibinfo{author}{D.~Guo}, \bibinfo{author}{Y.~Kim}, \bibinfo{author}{A.~Rush},
\newblock \bibinfo{title}{Sequence-{Level} {Mixed} {Sample} {Data} {Augmentation}},
\newblock in: \bibinfo{booktitle}{Proceedings of the 2020 {Conference} on {Empirical} {Methods} in {Natural} {Language} {Processing}}, EMNLP'20, \bibinfo{address}{Virtual Event}, \bibinfo{year}{2020}, pp. \bibinfo{pages}{5547--5552}. \URLprefix \url{https://aclanthology.org/2020.emnlp-main.447}. \DOIprefix\doi{10.18653/v1/2020.emnlp-main.447}.
\bibitem[{Clive et~al.(2022)Clive, Cao, and Rei}]{clive_control_2022}
\bibinfo{author}{J.~Clive}, \bibinfo{author}{K.~Cao}, \bibinfo{author}{M.~Rei},
\newblock \bibinfo{title}{Control {Prefixes} for {Parameter}-{Efficient} {Text} {Generation}}  (\bibinfo{year}{2022}). \URLprefix \url{http://arxiv.org/abs/2110.08329}. \DOIprefix\doi{10.48550/arXiv.2110.08329}, \bibinfo{note}{arXiv:2110.08329}.
\bibitem[{WebNLG(2024)}]{webnlg_papersdt_2024}
\bibinfo{author}{WebNLG}, \bibinfo{title}{{Papers} with {Code} - {WebNLG} {Dataset}}, \bibinfo{year}{2024}. \URLprefix \url{https://paperswithcode.com/dataset/webnlg}.
\bibitem[{Gardent et~al.(2017)Gardent, Shimorina, Narayan, and Perez-Beltrachini}]{gardent_webnlg_2017}
\bibinfo{author}{C.~Gardent}, \bibinfo{author}{A.~Shimorina}, \bibinfo{author}{S.~Narayan}, \bibinfo{author}{L.~Perez-Beltrachini},
\newblock \bibinfo{title}{The {WebNLG} {Challenge}: {Generating} {Text} from {RDF} {Data}},
\newblock in: \bibinfo{booktitle}{Proceedings of the 10th {International} {Conference} on {Natural} {Language} {Generation}}, INLG'17, \bibinfo{address}{Santiago de Compostela, Spain}, \bibinfo{year}{2017}, pp. \bibinfo{pages}{124--133}. \URLprefix \url{https://aclanthology.org/W17-3518}. \DOIprefix\doi{10.18653/v1/W17-3518}.
\bibitem[{Xu et~al.(2023)Xu, Zhu, Wang, and Zhang}]{xu_how_2023}
\bibinfo{author}{X.~Xu}, \bibinfo{author}{Y.~Zhu}, \bibinfo{author}{X.~Wang}, \bibinfo{author}{N.~Zhang},
\newblock \bibinfo{title}{How to {Unleash} the {Power} of {Large} {Language} {Models} for {Few}-shot {Relation} {Extraction}?},
\newblock in: \bibinfo{booktitle}{Proceedings of {The} {Fourth} {Workshop} on {Simple} and {Efficient} {Natural} {Language} {Processing}}, SustaiNLP'23, \bibinfo{address}{Toronto, Canada (Hybrid)}, \bibinfo{year}{2023}, pp. \bibinfo{pages}{190--200}. \URLprefix \url{https://aclanthology.org/2023.sustainlp-1.13}. \DOIprefix\doi{10.18653/v1/2023.sustainlp-1.13}.
\bibitem[{Ye et~al.(2022)Ye, Gao, Li, Xu, Feng, Wu, Yu, and Kong}]{ye_zerogen_2022}
\bibinfo{author}{J.~Ye}, \bibinfo{author}{J.~Gao}, \bibinfo{author}{Q.~Li}, \bibinfo{author}{H.~Xu}, \bibinfo{author}{J.~Feng}, \bibinfo{author}{Z.~Wu}, \bibinfo{author}{T.~Yu}, \bibinfo{author}{L.~Kong},
\newblock \bibinfo{title}{Zerogen: Efficient zero-shot learning via dataset generation},
\newblock in: \bibinfo{booktitle}{Proceedings of the 2022 Conference on Empirical Methods in Natural Language Processing}, EMNLP'22, \bibinfo{publisher}{Association for Computational Linguistics}, \bibinfo{year}{2022}, pp. \bibinfo{pages}{11653--11669}. \URLprefix \url{https://doi.org/10.18653/v1/2022.emnlp-main.801}. \DOIprefix\doi{10.18653/V1/2022.EMNLP-MAIN.801}.
\bibitem[{Thulasidasan et~al.(2019)Thulasidasan, Bhattacharya, Bilmes, Chennupati, and Mohd{-}Yusof}]{thulasidasan_combating_2019}
\bibinfo{author}{S.~Thulasidasan}, \bibinfo{author}{T.~Bhattacharya}, \bibinfo{author}{J.~A. Bilmes}, \bibinfo{author}{G.~Chennupati}, \bibinfo{author}{J.~Mohd{-}Yusof},
\newblock \bibinfo{title}{Combating label noise in deep learning using abstention},
\newblock in: \bibinfo{booktitle}{Proceedings of the 36th International Conference on Machine Learning}, volume~\bibinfo{volume}{97} of \textit{\bibinfo{series}{ICML'19}}, \bibinfo{publisher}{{PMLR}}, \bibinfo{year}{2019}, pp. \bibinfo{pages}{6234--6243}. \URLprefix \url{http://proceedings.mlr.press/v97/thulasidasan19a.html}.
\bibitem[{Ma et~al.(2020)Ma, Huang, Wang, Erfani, and Bailey}]{ma_normalized_2020}
\bibinfo{author}{X.~Ma}, \bibinfo{author}{H.~Huang}, \bibinfo{author}{Y.~Wang}, \bibinfo{author}{S.~R.~S. Erfani}, \bibinfo{author}{J.~Bailey},
\newblock \bibinfo{title}{Normalized loss functions for deep learning with noisy labels},
\newblock in: \bibinfo{booktitle}{Proceedings of the 37th {International} {Conference} on {Machine} {Learning}}, volume \bibinfo{volume}{119} of \textit{\bibinfo{series}{{ICML}'20}}, \bibinfo{year}{2020}, pp. \bibinfo{pages}{6543--6553}.
\bibitem[{Liu and Guo(2020)}]{liu_peer_2020}
\bibinfo{author}{Y.~Liu}, \bibinfo{author}{H.~Guo},
\newblock \bibinfo{title}{Peer loss functions: Learning from noisy labels without knowing noise rates},
\newblock in: \bibinfo{booktitle}{Proceedings of the 37th International Conference on Machine Learning}, volume \bibinfo{volume}{119} of \textit{\bibinfo{series}{ICML'20}}, \bibinfo{year}{2020}, pp. \bibinfo{pages}{6226--6236}. \URLprefix \url{http://proceedings.mlr.press/v119/liu20e.html}.
\bibitem[{Filippova(2020)}]{filippova_controlled_2020}
\bibinfo{author}{K.~Filippova},
\newblock \bibinfo{title}{Controlled {Hallucinations}: {Learning} to {Generate} {Faithfully} from {Noisy} {Data}},
\newblock in: \bibinfo{booktitle}{Findings of the 20210Conference on Empirical Methods in Natural Language Processing}, EMNLP'20, \bibinfo{address}{Online}, \bibinfo{year}{2020}, pp. \bibinfo{pages}{864--870}. \URLprefix \url{https://aclanthology.org/2020.findings-emnlp.76}. \DOIprefix\doi{10.18653/v1/2020.findings-emnlp.76}.
\bibitem[{Tan et~al.(2022)Tan, Xu, Bing, Ng, and Aljunied}]{tan_revisiting_2022}
\bibinfo{author}{Q.~Tan}, \bibinfo{author}{L.~Xu}, \bibinfo{author}{L.~Bing}, \bibinfo{author}{H.~T. Ng}, \bibinfo{author}{S.~M. Aljunied},
\newblock \bibinfo{title}{Revisiting {DocRED} - {Addressing} the {False} {Negative} {Problem} in {Relation} {Extraction}},
\newblock in: \bibinfo{booktitle}{Proceedings of the 2022 {Conference} on {Empirical} {Methods} in {Natural} {Language} {Processing}}, EMNLP'22, \bibinfo{address}{Abu Dhabi, United Arab Emirates}, \bibinfo{year}{2022}, pp. \bibinfo{pages}{8472--8487}. \URLprefix \url{https://aclanthology.org/2022.emnlp-main.580}. \DOIprefix\doi{10.18653/v1/2022.emnlp-main.580}.
\bibitem[{Stoica et~al.(2021)Stoica, Platanios, and Poczos}]{stoica_re-tacred_2021}
\bibinfo{author}{G.~Stoica}, \bibinfo{author}{E.~A. Platanios}, \bibinfo{author}{B.~Poczos},
\newblock \bibinfo{title}{Re-{TACRED}: {Addressing} {Shortcomings} of the {TACRED} {Dataset}},
\newblock \bibinfo{journal}{Proceedings of the AAAI Conference on Artificial Intelligence} \bibinfo{volume}{35} (\bibinfo{year}{2021}) \bibinfo{pages}{13843--13850}. \URLprefix \url{https://ojs.aaai.org/index.php/AAAI/article/view/17631}. \DOIprefix\doi{10.1609/aaai.v35i15.17631}.
\bibitem[{Yu et~al.(2019)Yu, Han, Yao, Niu, Tsang, and Sugiyama}]{yu_how_2019}
\bibinfo{author}{X.~Yu}, \bibinfo{author}{B.~Han}, \bibinfo{author}{J.~Yao}, \bibinfo{author}{G.~Niu}, \bibinfo{author}{I.~Tsang}, \bibinfo{author}{M.~Sugiyama},
\newblock \bibinfo{title}{How does {Disagreement} {Help} {Generalization} against {Label} {Corruption}?},
\newblock in: \bibinfo{booktitle}{Proceedings of the 36th {International} {Conference} on {Machine} {Learning}}, ICML'19, \bibinfo{year}{2019}, pp. \bibinfo{pages}{7164--7173}. \URLprefix \url{https://proceedings.mlr.press/v97/yu19b.html}.
\bibitem[{Ma et~al.(2018)Ma, Wang, Houle, Zhou, Erfani, Xia, Wijewickrema, and Bailey}]{ma_dimensionality-driven_2018}
\bibinfo{author}{X.~Ma}, \bibinfo{author}{Y.~Wang}, \bibinfo{author}{M.~E. Houle}, \bibinfo{author}{S.~Zhou}, \bibinfo{author}{S.~M. Erfani}, \bibinfo{author}{S.~Xia}, \bibinfo{author}{S.~N.~R. Wijewickrema}, \bibinfo{author}{J.~Bailey},
\newblock \bibinfo{title}{Dimensionality-driven learning with noisy labels},
\newblock in: \bibinfo{booktitle}{Proceedings of the 35th International Conference on Machine Learning}, volume~\bibinfo{volume}{80} of \textit{\bibinfo{series}{ICML'18}}, \bibinfo{year}{2018}, pp. \bibinfo{pages}{3361--3370}. \URLprefix \url{http://proceedings.mlr.press/v80/ma18d.html}.
\bibitem[{Reiter(2018)}]{reiter_structured_2018}
\bibinfo{author}{E.~Reiter},
\newblock \bibinfo{title}{A {Structured} {Review} of the {Validity} of {BLEU}},
\newblock \bibinfo{journal}{Computational Linguistics} \bibinfo{volume}{44} (\bibinfo{year}{2018}) \bibinfo{pages}{393--401}. \URLprefix \url{https://doi.org/10.1162/coli_a_00322}. \DOIprefix\doi{10.1162/coli_a_00322}.
\bibitem[{Falke et~al.(2019)Falke, Ribeiro, Utama, Dagan, and Gurevych}]{falke_ranking_2019}
\bibinfo{author}{T.~Falke}, \bibinfo{author}{L.~F.~R. Ribeiro}, \bibinfo{author}{P.~A. Utama}, \bibinfo{author}{I.~Dagan}, \bibinfo{author}{I.~Gurevych},
\newblock \bibinfo{title}{Ranking {Generated} {Summaries} by {Correctness}: {An} {Interesting} but {Challenging} {Application} for {Natural} {Language} {Inference}},
\newblock in: \bibinfo{booktitle}{Proceedings of the 57th {Annual} {Meeting} of the {Association} for {Computational} {Linguistics}}, ACL'19, \bibinfo{address}{Florence, Italy}, \bibinfo{year}{2019}, pp. \bibinfo{pages}{2214--2220}. \URLprefix \url{https://aclanthology.org/P19-1213}. \DOIprefix\doi{10.18653/v1/P19-1213}.
\bibitem[{Wang et~al.(2020)Wang, Wang, An, Yu, and Chen}]{wang_towards_2020}
\bibinfo{author}{Z.~Wang}, \bibinfo{author}{X.~Wang}, \bibinfo{author}{B.~An}, \bibinfo{author}{D.~Yu}, \bibinfo{author}{C.~Chen},
\newblock \bibinfo{title}{Towards {Faithful} {Neural} {Table}-to-{Text} {Generation} with {Content}-{Matching} {Constraints}},
\newblock in: \bibinfo{booktitle}{Proceedings of the 58th {Annual} {Meeting} of the {Association} for {Computational} {Linguistics}}, ACL'20, \bibinfo{year}{2020}, pp. \bibinfo{pages}{1072--1086}. \URLprefix \url{http://arxiv.org/abs/2005.00969}. \DOIprefix\doi{10.18653/v1/2020.acl-main.101}, \bibinfo{note}{arXiv:2005.00969}.
\bibitem[{Shuster et~al.(2021)Shuster, Poff, Chen, Kiela, and Weston}]{shuster_retrieval_2021}
\bibinfo{author}{K.~Shuster}, \bibinfo{author}{S.~Poff}, \bibinfo{author}{M.~Chen}, \bibinfo{author}{D.~Kiela}, \bibinfo{author}{J.~Weston},
\newblock \bibinfo{title}{Retrieval {Augmentation} {Reduces} {Hallucination} in {Conversation}},
\newblock in: \bibinfo{booktitle}{Findings of the 2021 Conference on Empirical Methods in Natural Language Processing}, EMNLP'21, \bibinfo{address}{Punta Cana, Dominican Republic}, \bibinfo{year}{2021}, pp. \bibinfo{pages}{3784--3803}. \URLprefix \url{https://aclanthology.org/2021.findings-emnlp.320}. \DOIprefix\doi{10.18653/v1/2021.findings-emnlp.320}.
\bibitem[{Martindale et~al.(2019)Martindale, Carpuat, Duh, and McNamee}]{martindale_identifying_2019}
\bibinfo{author}{M.~Martindale}, \bibinfo{author}{M.~Carpuat}, \bibinfo{author}{K.~Duh}, \bibinfo{author}{P.~McNamee},
\newblock \bibinfo{title}{Identifying {Fluently} {Inadequate} {Output} in {Neural} and {Statistical} {Machine} {Translation}},
\newblock in: \bibinfo{booktitle}{Proceedings of {Machine} {Translation} {Summit} {XVII}}, \bibinfo{address}{Dublin, Ireland}, \bibinfo{year}{2019}, pp. \bibinfo{pages}{233--243}. \URLprefix \url{https://aclanthology.org/W19-6623}.
\bibitem[{Liu et~al.(2021)Liu, Zheng, Chang, and Sui}]{liu_towards_2021}
\bibinfo{author}{T.~Liu}, \bibinfo{author}{X.~Zheng}, \bibinfo{author}{B.~Chang}, \bibinfo{author}{Z.~Sui},
\newblock \bibinfo{title}{Towards {Faithfulness} in {Open} {Domain} {Table}-to-text {Generation} from an {Entity}-centric {View}},
\newblock volume~\bibinfo{volume}{35} of \textit{\bibinfo{series}{AAAI'21}}, \bibinfo{year}{2021}, pp. \bibinfo{pages}{13415--13423}. \URLprefix \url{https://ojs.aaai.org/index.php/AAAI/article/view/17583}. \DOIprefix\doi{10.1609/aaai.v35i15.17583}.
\bibitem[{Dušek and Kasner(2020)}]{dusek_evaluating_2020}
\bibinfo{author}{O.~Dušek}, \bibinfo{author}{Z.~Kasner},
\newblock \bibinfo{title}{Evaluating {Semantic} {Accuracy} of {Data}-to-{Text} {Generation} with {Natural} {Language} {Inference}},
\newblock in: \bibinfo{booktitle}{Proceedings of the 13th {International} {Conference} on {Natural} {Language} {Generation}}, ICNLG'20, \bibinfo{address}{Dublin, Ireland}, \bibinfo{year}{2020}, pp. \bibinfo{pages}{131--137}. \URLprefix \url{https://aclanthology.org/2020.inlg-1.19}. \DOIprefix\doi{10.18653/v1/2020.inlg-1.19}.
\bibitem[{Tian et~al.(2020)Tian, Narayan, Sellam, and Parikh}]{tian_sticking_2020}
\bibinfo{author}{R.~Tian}, \bibinfo{author}{S.~Narayan}, \bibinfo{author}{T.~Sellam}, \bibinfo{author}{A.~P. Parikh}, \bibinfo{title}{Sticking to the {Facts}: {Confident} {Decoding} for {Faithful} {Data}-to-{Text} {Generation}}, \bibinfo{year}{2020}. \URLprefix \url{http://arxiv.org/abs/1910.08684}. \DOIprefix\doi{10.48550/arXiv.1910.08684}, \bibinfo{note}{arXiv:1910.08684 [cs]}.
\bibitem[{Nie et~al.(2019)Nie, Yao, Wang, Pan, and Lin}]{nie_simple_2019}
\bibinfo{author}{F.~Nie}, \bibinfo{author}{J.-G. Yao}, \bibinfo{author}{J.~Wang}, \bibinfo{author}{R.~Pan}, \bibinfo{author}{C.-Y. Lin},
\newblock \bibinfo{title}{A {Simple} {Recipe} towards {Reducing} {Hallucination} in {Neural} {Surface} {Realisation}},
\newblock in: \bibinfo{booktitle}{Proceedings of the 57th {Annual} {Meeting} of the {Association} for {Computational} {Linguistics}}, ACL'19, \bibinfo{address}{Florence, Italy}, \bibinfo{year}{2019}, pp. \bibinfo{pages}{2673--2679}. \URLprefix \url{https://aclanthology.org/P19-1256}. \DOIprefix\doi{10.18653/v1/P19-1256}.
\bibitem[{Yan et~al.(2021)Yan, Zhang, Fu, Zhang, and Wei}]{yan_partition_2021}
\bibinfo{author}{Z.~Yan}, \bibinfo{author}{C.~Zhang}, \bibinfo{author}{J.~Fu}, \bibinfo{author}{Q.~Zhang}, \bibinfo{author}{Z.~Wei},
\newblock \bibinfo{title}{A {Partition} {Filter} {Network} for {Joint} {Entity} and {Relation} {Extraction}},
\newblock in: \bibinfo{booktitle}{Proceedings of the 2021 {Conference} on {Empirical} {Methods} in {Natural} {Language} {Processing}}, EMNLP'21, \bibinfo{address}{Online and Punta Cana, Dominican Republic}, \bibinfo{year}{2021}, pp. \bibinfo{pages}{185--197}. \URLprefix \url{https://aclanthology.org/2021.emnlp-main.17}. \DOIprefix\doi{10.18653/v1/2021.emnlp-main.17}.
\bibitem[{Yao et~al.(2019)Yao, Ye, Li, Han, Lin, Liu, Liu, Huang, Zhou, and Sun}]{yao_docred_2019}
\bibinfo{author}{Y.~Yao}, \bibinfo{author}{D.~Ye}, \bibinfo{author}{P.~Li}, \bibinfo{author}{X.~Han}, \bibinfo{author}{Y.~Lin}, \bibinfo{author}{Z.~Liu}, \bibinfo{author}{Z.~Liu}, \bibinfo{author}{L.~Huang}, \bibinfo{author}{J.~Zhou}, \bibinfo{author}{M.~Sun},
\newblock \bibinfo{title}{{DocRED}: {A} {Large}-{Scale} {Document}-{Level} {Relation} {Extraction} {Dataset}},
\newblock in: \bibinfo{booktitle}{Proceedings of the 57th {Annual} {Meeting} of the {Association} for {Computational} {Linguistics}}, ACL'19, \bibinfo{address}{Florence, Italy}, \bibinfo{year}{2019}, pp. \bibinfo{pages}{764--777}. \URLprefix \url{https://aclanthology.org/P19-1074}. \DOIprefix\doi{10.18653/v1/P19-1074}.
\bibitem[{Ma et~al.(2023)Ma, Wang, and Okazaki}]{ma_dreeam_2023}
\bibinfo{author}{Y.~Ma}, \bibinfo{author}{A.~Wang}, \bibinfo{author}{N.~Okazaki},
\newblock \bibinfo{title}{{DREEAM}: {Guiding} {Attention} with {Evidence} for {Improving} {Document}-{Level} {Relation} {Extraction}}  (\bibinfo{year}{2023}). \URLprefix \url{http://arxiv.org/abs/2302.08675}. \DOIprefix\doi{10.48550/arXiv.2302.08675}, \bibinfo{note}{arXiv:2302.08675}.
\bibitem[{doc(2024)}]{docred_papers_2024}
\bibinfo{title}{Papers with {Code} - {DocRED} {Benchmark} ({Relation} {Extraction})}, \bibinfo{year}{2024}. \URLprefix \url{https://paperswithcode.com/sota/relation-extraction-on-docred}.
\bibitem[{red(2024)}]{redocred_papers_2024}
\bibinfo{title}{Papers with {Code} - {ReDocRED} {Benchmark} ({Relation} {Extraction})}, \bibinfo{year}{2024}. \URLprefix \url{https://paperswithcode.com/sota/relation-extraction-on-redocred}.
\bibitem[{Aghajanyan et~al.(2022)Aghajanyan, Okhonko, Lewis, Joshi, Xu, Ghosh, and Zettlemoyer}]{aghajanyan_htlm_2021}
\bibinfo{author}{A.~Aghajanyan}, \bibinfo{author}{D.~Okhonko}, \bibinfo{author}{M.~Lewis}, \bibinfo{author}{M.~Joshi}, \bibinfo{author}{H.~Xu}, \bibinfo{author}{G.~Ghosh}, \bibinfo{author}{L.~Zettlemoyer},
\newblock \bibinfo{title}{{HTLM:} hyper-text pre-training and prompting of language models},
\newblock in: \bibinfo{booktitle}{Proceedings of the 10th International Conference on Learning Representations}, ICLR'22, \bibinfo{year}{2022}.
\bibitem[{Zeng et~al.(2018)Zeng, Zeng, He, Liu, and Zhao}]{zeng_extracting_2018}
\bibinfo{author}{X.~Zeng}, \bibinfo{author}{D.~Zeng}, \bibinfo{author}{S.~He}, \bibinfo{author}{K.~Liu}, \bibinfo{author}{J.~Zhao},
\newblock \bibinfo{title}{Extracting {Relational} {Facts} by an {End}-to-{End} {Neural} {Model} with {Copy} {Mechanism}},
\newblock in: \bibinfo{booktitle}{Proceedings of the 56th {Annual} {Meeting} of the {Association} for {Computational} {Linguistics}}, ACL'18, \bibinfo{address}{Melbourne, Australia}, \bibinfo{year}{2018}, pp. \bibinfo{pages}{506--514}. \URLprefix \url{https://aclanthology.org/P18-1047}. \DOIprefix\doi{10.18653/v1/P18-1047}.
\bibitem[{WebNLG(2024)}]{webnlg_papersre_2024}
\bibinfo{author}{WebNLG}, \bibinfo{title}{{PapersRE} with {Code} - {WebNLG} {Benchmark} ({Relation} {Extraction})}, \bibinfo{year}{2024}. \URLprefix \url{https://paperswithcode.com/sota/relation-extraction-on-webnlg}, \bibinfo{note}{available at https://paperswithcode.com/sota/relation-extraction-on-webnlg}.
\bibitem[{Touvron et~al.(2023)Touvron, Martin et~al.}]{touvron_llama_2023}
\bibinfo{author}{H.~Touvron}, \bibinfo{author}{L.~Martin}, et~al.,
\newblock \bibinfo{title}{Llama 2: Open foundation and fine-tuned chat models},
\newblock \bibinfo{journal}{CoRR} \bibinfo{volume}{abs/2307.09288} (\bibinfo{year}{2023}). \URLprefix \url{https://doi.org/10.48550/arXiv.2307.09288}. \DOIprefix\doi{10.48550/ARXIV.2307.09288}. \href{http://arxiv.org/abs/2307.09288}{{\tt arXiv:2307.09288}}.
\bibitem[{Gardent et~al.(2017)Gardent, Shimorina, Narayan, and Perez-Beltrachini}]{gardent_creating_2017}
\bibinfo{author}{C.~Gardent}, \bibinfo{author}{A.~Shimorina}, \bibinfo{author}{S.~Narayan}, \bibinfo{author}{L.~Perez-Beltrachini},
\newblock \bibinfo{title}{Creating {Training} {Corpora} for {NLG} {Micro}-{Planners}},
\newblock in: \bibinfo{editor}{R.~Barzilay}, \bibinfo{editor}{M.-Y. Kan} (Eds.), \bibinfo{booktitle}{Proceedings of the 55th {Annual} {Meeting} of the {Association} for {Computational} {Linguistics} ({Volume} 1: {Long} {Papers})}, \bibinfo{publisher}{Association for Computational Linguistics}, \bibinfo{address}{Vancouver, Canada}, \bibinfo{year}{2017}, pp. \bibinfo{pages}{179--188}. \URLprefix \url{https://aclanthology.org/P17-1017}. \DOIprefix\doi{10.18653/v1/P17-1017}.
\bibitem[{noa(2024)}]{noauthor_sentence-transformersall-mpnet-base-v2_2024}
\bibinfo{title}{sentence-transformers/all-mpnet-base-v2 · {Hugging} {Face}}, \bibinfo{year}{2024}. \URLprefix \url{https://huggingface.co/sentence-transformers/all-mpnet-base-v2}.
\bibitem[{Liu et~al.(2019)Liu, Ott, Goyal, Du, Joshi, Chen, Levy, Lewis, Zettlemoyer, and Stoyanov}]{liu_roberta_2019}
\bibinfo{author}{Y.~Liu}, \bibinfo{author}{M.~Ott}, \bibinfo{author}{N.~Goyal}, \bibinfo{author}{J.~Du}, \bibinfo{author}{M.~Joshi}, \bibinfo{author}{D.~Chen}, \bibinfo{author}{O.~Levy}, \bibinfo{author}{M.~Lewis}, \bibinfo{author}{L.~Zettlemoyer}, \bibinfo{author}{V.~Stoyanov},
\newblock \bibinfo{title}{{RoBERTa}: {A} {Robustly} {Optimized} {BERT} {Pretraining} {Approach}}  (\bibinfo{year}{2019}). \URLprefix \url{http://arxiv.org/abs/1907.11692}. \DOIprefix\doi{10.48550/arXiv.1907.11692}, \bibinfo{note}{arXiv:1907.11692}.
\bibitem[{He et~al.(2021)He, Liu, Gao, and Chen}]{he_deberta_2021}
\bibinfo{author}{P.~He}, \bibinfo{author}{X.~Liu}, \bibinfo{author}{J.~Gao}, \bibinfo{author}{W.~Chen},
\newblock \bibinfo{title}{Deberta: decoding-enhanced bert with disentangled attention},
\newblock in: \bibinfo{booktitle}{Proceedings of the 9th International Conference on Learning Representations}, ICLR'21, \bibinfo{year}{2021}. \URLprefix \url{https://openreview.net/forum?id=XPZIaotutsD}.

\end{thebibliography}



\end{document}